\documentclass[letterpaper, 10 pt, conference]{ieeeconf}  

\IEEEoverridecommandlockouts                              

\usepackage{times}

\usepackage{multicol}
\usepackage{multirow}
\usepackage{hyperref}
\hypersetup{bookmarks=true}
\usepackage{amsmath}
\usepackage{amssymb}
\usepackage{xcolor}
\usepackage{optidef}
\usepackage{booktabs}
\usepackage{graphicx}
\usepackage{bm}
\let\labelindent\relax 
\usepackage{enumitem}
\usepackage{todonotes}
\usepackage{lipsum}
\usepackage{capt-of,etoolbox}
\usepackage{mwe}
\usepackage{svg}
\usepackage[linesnumbered,ruled,vlined]{algorithm2e}
\usepackage[dvipsnames]{xcolor}
\usepackage[table]{xcolor}

\renewcommand{\bold}[1]{\mathbf{#1}} 
\newcommand{\R}[0]{\mathbb{R}} 

\newcommand{\numHydros}[0]{M}
\newcommand{\numContacts}[0]{N}

\newcommand{\sethree}[0]{\text{SE}(3)}
\newcommand{\mass}[0]{m}
\newcommand{\fricCoeff}[0]{\mu}

\newcommand{\externalWrench}[0]{\mathbf{w}_{\text{ext}}}

\newcommand{\frictionCone}[1]{\mathcal K^{3}_{#1}}

\newcommand{\object}[1]{{#1}^\text{O}}
\newcommand{\hydro}[1]{{#1}^\text{H}}
\newcommand{\pose}[0]{\bold q}
\newcommand{\objectPose}[1]{\object{\pose}_{#1}}
\newcommand{\hydroPose}[1]{\hydro{\pose}_{#1}}
\newcommand{\dobjectPose}[1]{\object{\dot{\pose}}_{#1}}
\newcommand{\dhydroPose}[1]{\hydro{\dot{\pose}}_{#1}}
\newcommand{\ddobjectPose}[1]{\object{\ddot{\pose}}_{#1}}
\newcommand{\ddhydroPose}[1]{\hydro{\ddot{\pose}}_{#1}}

\newcommand{\contactForce}[1]{\bold f^{\text{c}_{#1}}}
\newcommand{\dcontactForce}[1]{\dot{\bold f}^{\text{c}_{#1}}}
\newcommand{\resetcontactForce}[1]{\tilde{\bold f}^{\text{c}_{#1}}}
\newcommand{\projectcontactForce}[1]{\bar{\bold f}^{\text{c}_{#1}}}
\newcommand{\contactForces}[1]{\bold F^{\text{c}_{#1}}}

\newcommand{\contactJacobianObj}[1]{^{\text{O}} \bold J_{#1}^{\text{c}_i}}
\newcommand{\contactJacobianHydro}[1]{^{\text{H}} \bold J_{#1}^{\text{c}_i}}

\newcommand{\inertiaMatrix}[0]{\bold M}
\newcommand{\taug}[0]{\boldsymbol \tau_g}

\newcommand{\state}[1]{\bold x_\text{#1}}
\newcommand{\action}[1]{\bold u_\text{#1}}

\newcommand{\sdf}[0]{\phi}
\newcommand{\relu}[0]{\text{ReLU}}

\newcommand{\objectPoseGoal}[1]{\objectPose{\text d}}

\newcommand{\dc}[0]{\cellcolor{gray!20}} 

\title{\LARGE \bf
Hydrosoft: Non-Holonomic Hydroelastic Models for Compliant Tactile Manipulation
}

\author{
Miquel Oller\hspace{15pt} An Dang \hspace{15pt} Nima Fazeli\\
Department of Robotics, University of Michigan \\
Ann Arbor, MI 48109, United States\\
\texttt{\{oller, andang, nfz\}@umich.edu} \\
\url{https://www.mmintlab.com/hydrosoft} 
}

\begin{document}



\makeatletter
\def\@maketitle{%
  \newpage
  \null
  \vskip 2em%
  \begin{center}%
    {\LARGE \@title \par}%
    \vskip 1.5em%
    {\large \@author \par}%
  \end{center}%
  \vskip 1em
  \begin{center}
    \includegraphics[width=\linewidth]{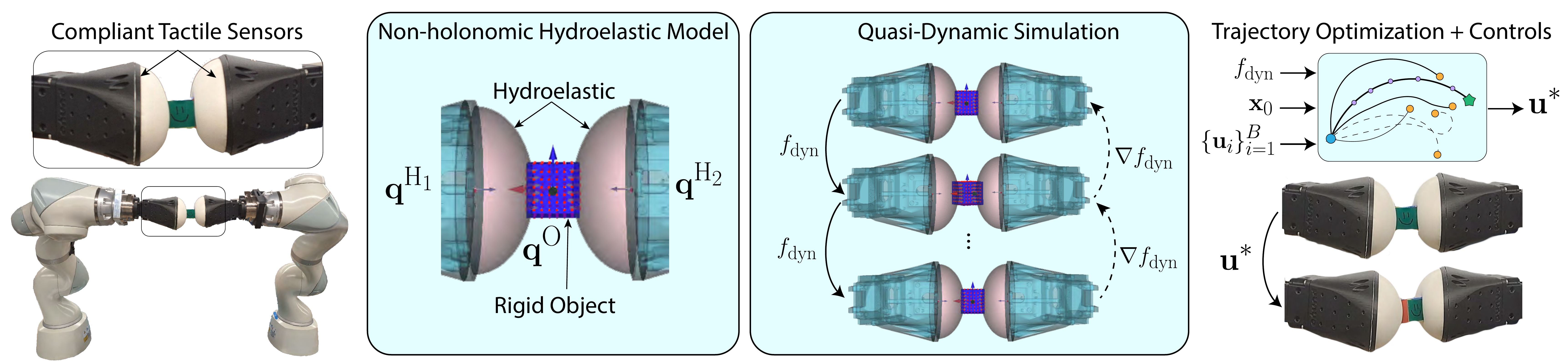}
  \end{center}
  \setcounter{figure}{0}
  \captionof{figure}{\textbf{Approach Overview:} \textit{First panel:} The robot grasps an object using compliant tactile sensors and performs in-hand rotation. \textit{Second panel:} Our method models the tactile sensors as non-holonomic hydroelastic elements. \textit{Third panel:} The non-holonomic hydroelastic model is integrated with differentiable dynamics. \textit{Fourth panel:} The resulting dynamics are used with gradient-based trajectory optimization to control and execute the task.}

  \vskip 1em
}
\makeatother


\maketitle


\begin{abstract}

Tactile sensors have long been valued for their perceptual capabilities, offering rich insights into the otherwise hidden interface between the robot and grasped objects. Yet their inherent compliance—a key driver of force-rich interactions—remains underexplored. The central challenge is to capture the complex, nonlinear dynamics introduced by these passive compliant elements. Here, we present a computationally efficient non-holonomic hydroelastic model that accurately models path-dependent contact force distributions and dynamic surface area variations. Our insight is to extend the object’s state space, explicitly incorporating the distributed forces generated by the compliant sensor. Our differentiable formulation not only accounts for path-dependent behavior but also enables gradient-based trajectory optimization, seamlessly integrating with high-resolution tactile feedback. We demonstrate the effectiveness of our approach across a range of simulated and real-world experiments and demonstrate the importance of modeling the path dependence of sensor dynamics.

\end{abstract}


\section{Introduction}
\label{sec:introduction}

Contact-rich object manipulation is a cornerstone of dexterous robotic systems, enabling precise control over an object’s position and orientation by regulating the forces at the robot–object interface \cite{mason_manipulation, hogan_impedance_control}. A particularly effective way to achieve this regulation is through mechanical compliance at the point of contact \cite{pratt1995series, manipulation_via_membranes, seed, tactile_nonprehensile}, which can absorb uncertainty and maintain stable force exchange in complex manipulation tasks. 

However, capturing the nonlinear, high-dimensional dynamics of compliant end-effectors remains a open problem. On one end of the spectrum, finite element methods can accurately model deformation but are often computationally prohibitive for real-time control \cite{bubble_fem}. On the other, simplified spring-damper approximations offer speed but fail to generalize to tasks where the distribution of contact forces and surface deformations play a pivotal role. To bridge this gap, hydroelastic modeling has emerged as an appealing middle ground, balancing physical fidelity with computational efficiency \cite{pressure_field, bubbles_hydroelastic}. By focusing on force distributions and contact-surface variations, hydroelastic models bypass much of the complexity associated with fully simulating elastic deformations—making them especially well suited to contact-rich manipulation.

In this work, we propose Hydrosoft: a non-honolonomic extension to the hydroelastic model that models: (i) the path dependence of force build up in the sensor substrate; (ii) Coulomb friction; (iii) non-smooth contact transitions. Our approach incorporates relaxations for smooth gradients, offers efficient and parallelized computation, and supports high-resolution tactile sensing for real-time feedback and control. By incrementally updating the contact forces, our method models the history-dependent deformation and force build up of the sensor substrate while preserving the computational benefits of hydroelastic models. As a result, our approach enables gradient-based planning and control in scenarios that demand both precise force regulation and continuous contact ranging from rolling objects on a table to in-hand object reconfiguration.

In more detail, our model addresses 4 key limitations of prior models: First, the original formulation \cite{pressure_field} does not model static friction forces. This is of critical importance when considering grasping and object transitions in the grasp. Second, the hydroelastic formulation does not explicitly model deformation. As a result, sticking shear deformations may result in breaking contact locally between the hydroelastic and rigid objects. Intuitively, when a deformable is in contact and shear is applied, the interface can stick and due to compliance, the contact is preserved. However, this important effect is currently not modeled. Third, the contact forces are solely a function of the relative position between the objects, without accounting for the prior trajectory. Consequently, the computation of contact forces assumes path independence. However, this assumption breaks down in cases involving shear deformations, where the deformation history significantly influences the resulting forces. Finally, the hybrid nature of contact results in the model exhibiting non-smooth dynamics. As a result, the model is only able to produce sub-gradients. Specifically, only the differential elements within the hydroelastic contact region contribute to the gradients, while elements outside the contact region have zero gradients. This limitation is particularly evident in scenarios such as in-hand object rolling, where smooth transitions in contact are critical for accurate force and motion control.

\section{Related Work}
\label{sec:related_work}

\subsection{Rigid Body Dynamics with Frictional Contacts}
Modeling contact dynamics is challenging due to its hybrid nature, which leads to formulations with complementarity constraints to enforce non-penetration and Coulomb friction. Such problems are often posed as Nonlinear Complementarity Problems (NCPs), but can be approximated by Mixed Linear Complementarity Problems (MLCPs) through polyhedral friction cone approximations \cite{anitescu_potra, stewart_trinkle}. Anitescu et al. \cite{anitescu2006optimization} reformulated the problem as a Second-Order Cone Program (SOCP), interpretable as KKT conditions of a convex program, enabling gradient propagation \cite{pang_quasistatic}. Alternative formulations avoid complementarity: \cite{todorov_implicit} proposed an unconstrained optimization using intermediate variables, while \cite{complementarity_free_contact} introduced a differentiable closed-form model with spring forces in the contact dual cone. However, these works mainly consider rigid point contacts, whereas our focus is on elastic elements where deformation alleviates interpenetration issues.

\subsection{Smoothing and Differentiability}
Contact dynamics are also non-smooth, restricting optimization to sub-gradients. Smoothing methods address this: randomized smoothing \cite{suh_randomized_smoothing}, log-barrier smoothing \cite{pang2023global, dojo}, and $\texttt{softplus}$-based relaxations \cite{complementarity_free_contact}. These approaches enable differentiability but may introduce artifacts such as forces-at-a-distance. We similarly apply smoothing to support gradient-based trajectory optimization, while incorporating contact feedback to mitigate model mismatch.

\subsection{Modeling Patch Contact}

Patch contact has been studied via sliding models \cite{howe_cutkosky_sliding, limit_surface}, extended to precise manipulation \cite{xili_dual_limit_surface, bimbo_sliding}, though often limited to circular patches and uniform pressure assumptions. Composite friction cones approximate polygonal patches with edge-based cones \cite{manipulation_shared_grasping}, naturally capturing transitions from patch to line or point contacts. 
Patch contact has also been modeled for deformable elements, with approaches that estimate contact regions from tactile deformation \cite{bubble_contact_estimation} or jointly recover both deformations and forces \cite{nisp, bubble_fem}. Despite the physical realism, they are often computationally demanding, limiting their use in real-time manipulation. 

Hydroelastic models \cite{pressure_field, bubbles_hydroelastic} provide an alternative, where patch size emerges from geometry intersections. The Pressure Field Contact (PFC) model \cite{pressure_field} is efficient but lacks static forces; \cite{bubbles_hydroelastic} extended it to include static friction via optimization, later combined with iLQR for manipulation and locomotion \cite{hyroelastic_ilqr}, though at high computational cost. We build on hydroelastic formulations for fast, efficient modeling of highly elastic elements to regularize contact size in manipulation.

Contact patch modeling is of special interest in tactile simulation. Differentiable frameworks \cite{tactile_sim} and libraries like TacSL \cite{tacsl} enable realistic visuotactile outputs, but they often fail to capture tangential forces under sticking contact. In contrast, our formulation produces tangential forces for static friction consistent with Coulomb’s model, improving realism in contact-rich interactions.



\section{Problem Statement}
\label{sec:problem_statement}

We consider the problem of manipulating a rigid object with known geometry and physical parameters from an initial configuration $\objectPose{0}$ to a desired goal configuration $\objectPoseGoal{}$. 
The robot is equipped with $N$ passive-compliant elements, instantiated here as tactile sensors, which are rigidly attached to it. The robot has full control over the $\mathbb{SE}(3)$ poses of the attachment points, and thus the motion of the compliant elements is implicitly determined by the robot’s configuration.
The objective is to compute a sequence of attachment-point motions ${\hydroPose{1}, \dots, \hydroPose{K}}$ that drive the object toward the goal configuration $\objectPoseGoal{}$ while respecting the physical constraints arising from the interaction between the robot, the compliant elements, the object, and the environment.



\begin{table}[]
    \centering
    \begin{tabular}{ccc}
    \toprule
    Symbol & Space & Description \\ 
    \cmidrule(lr){1-1} \cmidrule(lr){2-2} \cmidrule(lr){3-3}
      $\mass$ &  $\R_{+}$ & Object Mass \\
      $\fricCoeff$ & $\R_{+}$ & Contact Friction \\
      $\mathbf{w}_\text{ext}$ & $\R^6$ & External Wrench \\
      $\objectPose{}$ & $\sethree$ & Rigid Object Pose \\
      $\hydroPose{}$ & $\sethree$ & Hydroelastic Object Pose \\
      $\objectPoseGoal{}$ & $\sethree$ & Desired Rigid Object Pose \\
      $\contactForce{}$ & $\frictionCone{\fricCoeff}$  & Contact Force \\ 
      $\contactForces{}$ & $\R^{\numContacts\times \numHydros\times 3}$ & Contact Forces \\
      $\contactJacobianObj{}$ & $\R^{6\times 3}$ & Contact Jacobian \\
      $\inertiaMatrix$ & $\R^{6\times 6}$ & Generalized Inertia Matrix \\
    \bottomrule
    \end{tabular}
    \label{tab:notation}
    \captionof{table}{\textbf{Notation}}
    \vspace*{-25 pt}
\end{table}


\section{Methodology}
\label{sec:methods}



Our methodology combines hydroelastic modeling of compliant contact with trajectory optimization to enable dexterous manipulation. The section is organized as follows.

First, we introduce the hydroelastic formulation used to resolve compliant contact forces between the object and the robot’s tactile elements (Sec. \ref{sec:methods:hydro_model}). We then extend this formulation to a non-holonomic incremental model that captures path-dependent deformations and frictional effects (Sec. \ref{sec:methods:model}). To improve computational efficiency and differentiability, we further propose a smoothed variant of the hydroelastic model (Sec. \ref{sec:methods:smoothed_hydro_model}).
Next, we embed the contact model into a quasi-dynamic system representation that relates robot control actions—changes in the attachment poses of the compliant elements—to object motion (Sec. \ref{sec:methods:quasidyn_formulation}). Finally, we formulate the trajectory optimization problem as a model predictive control (MPC), which computes the sequence of robot actions that achieves the desired object configuration (Sec. \ref{sec:methods:control}).


\begin{figure}
    \centering
    \includegraphics[width=\linewidth]{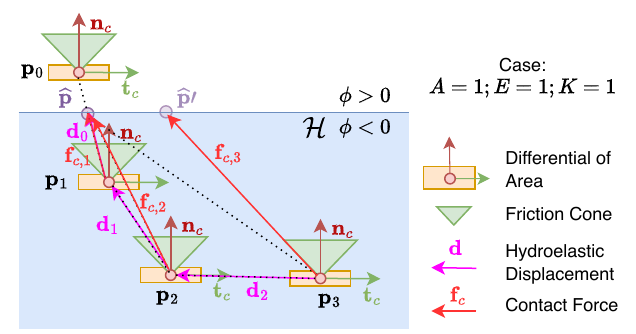}
    \caption{\textbf{Nonholonomic Hydroelastic Virtual Stick-Slip Behaviour:} We show the transition of a differential of the area along the hydroelastic body $\mathcal H$. Starting at a contact-free position $\bold p_0$, i.e. $\phi(\bold p_0)>0$, transition 0-1 shows a contact transition where the contact force is proportional to the displacement within the hydroelastic body $\bold d_0$. Note that this creates a virtual reference point $\hat{\bold p}$ at the boundary of the hydroelastic body. $\hat{\bold p}$ can be interpreted as the point corresponding to zero contact force in a slip-free transition. Transition 1-2 show a 'sticking' transition within the hydroelastic element. Note that contact force stays within the friction cone and the virtual reference point remains the same. Finally, transition 2-3 corresponds to a 'slipping' transition, where Coulumb's friction constraints make the boundary reference point $\hat{\bold p}$ has virtually translated along the hydroelastic boundary to $\hat{\bold p}'$}
    \label{fig:hydro_virtual_slip_stick}
    \vspace{-15pt}
\end{figure}

\subsection{Compliant Contact Forces via Hydroelastic Models}
\label{sec:methods:hydro_model}
We will start by introducing the hydroelastic formulation proposed in \cite{pressure_field} and \cite{bubbles_hydroelastic} adapted for our problem formulated in section \ref{sec:problem_statement}. We will consider the case of resolving the contact forces for an interaction between a rigid object $\text{O}$ and a hydroelastic body $\text{H}$. In section \ref{sec:methods:model} we will extend the formulation to the non-holonomic case.

The rigid object $\text{O}$ is discretized into a set of surface points $\mathcal O$ where 

\begin{equation}
\mathcal O = \{\bold p_1, \dots, \bold p_N \ | \ \bold p_i \in \R^3\}
\end{equation}
Each point $\bold p_i$ has associated a surface area $A_i$ and a contact frame centered at the point. We denote as $^\text{O}\bold q^{c_i} $ as the configuration of the contact frame $i \in \{1,\dots, N\}$ w.r.t the object frame.

We denote $\sdf$ as the signed-distance function (SDF) of the hydroelastic body for its nominal shape, i.e. contact-free shape. Therefore, we can define the hydroelastic domain as all the points inside the hydroelastic element:
\begin{equation}
    \mathcal H \triangleq \{\bold p \in \R^3 | \phi(\bold p) \leq 0\}
\end{equation}

We seek to compute the contact force associated to each surface point $\bold p_i \in \mathcal O$, denoted as $\contactForce{i}$. 
The normal contact force is proportional to the area $A_i$ associated to that point and the penetration distance on the hydroelastic body. The penetration distance is computed using the SDF function. This takes the following expression:

\begin{equation}
    f_{n,i} = \max(-EA_i\phi(\bold p_i), 0)
\end{equation}
where $E>0$ is the effective pressure gradient (in $\text{[Pa/m]}$).

To compute the tangential contact forces $\contactForce{i}_t$, \cite{pressure_field} assumes them to be $0$ if there is no relative motion between the contact point $\bold p_i$ and the hydroelastic body $\text{H}$ in the tangent plane, i.e. $^{\text{O}}\dot{\bold q}^{\text{c}_i}_{\perp} = \bold 0$. Otherwise, it follows Coulumb's friction.

\begin{equation}
    \contactForce{i}_t = \begin{cases} 
        \bold 0\in\R^3 & \text{if } \|^{\text{O}}\dot{\bold q}^{\text{c}_i}_{\perp}\| = 0\\
        -\mu f_{n,i} \frac{^{\text{O}}\dot{\bold q}^{\text{c}_i}_{\perp}}{\| ^{\text{O}}\dot{\bold q}^{\text{c}_i}_{\perp}\|}& \text{if } \|^{\text{O}}\dot{\bold q}^{\text{c}_i}_{\perp}\| > 0\\
    \end{cases}
\end{equation}

Similarly, \cite{bubbles_hydroelastic} computes the friction force using complimentary conditions derived from the maximum dissipation principle. Note that this constraint is combined with other contact constraints and dynamic constraints to resolve the contact via optimization.

Finally, the contact force is computed by combining normal and tangential forces
\begin{equation}
    \contactForce{i} = f_{n,i}\hat{\bold n} + \contactForce{i}_t
\end{equation}

\subsection{Non-holonomic Incremental Hydroelastic Model}
 \label{sec:methods:model}

Our goal is to derive a model capable of computing the reaction forces resulting from the contact interaction between a hydroelastic body and a rigid object. Our main considerations are the deformation path, surface area variations, and obeying Coulomb's friction. To this end, we extend the hydroelastic formulation from \cite{pressure_field}. Here, we derive the equations for a single hydroelastic body, but they can trivially be extended to consider $\numHydros$ bodies.


We start off with the standard Equations of Motion (EoM) for our system:
\begin{align}\label{eq:EoM-basic}
\bold M_\text{H}(\hydroPose{})\ddhydroPose{} + \bold C_\text{H}(\hydroPose{}, \dhydroPose{})\dhydroPose{} &= \bold \tau^\text{H}_g + \externalWrench^\text{H} + \action{} \\
\bold M_\text{O}(\objectPose{})\ddobjectPose{} + \bold C_\text{O}(\objectPose{}, \dobjectPose{})\dobjectPose{} &= \bold \tau^\text{O}_g + \externalWrench^\text{O}
\end{align}
where $\externalWrench^\text{H}$ and $\externalWrench^\text{O}$ are external wrenches computed from our hydroelastic contact model, $\bold M$ denotes the inertia matrix, $\bold C$ denotes the Coriolis and Centrifugal forces, $\bold \tau_g$ the torques induced by gravity, and $\action{}$ denotes the control input.


In our hydroelastic formulation, we individually compute the contact forces $\contactForce{i}$ for each contact point on the rigid object $\bold p_i$. 
Instead of calculating reaction forces through inter-object penetration from the closest boundary point, 
 we calculate the reaction forces by tracking the in-penetration displacements. This formulation approximates membrane deformation over time and provides a more accurate model for modelling the deformable object dynamics without accruing significant computational cost.


We denote $E$ and $K$ as the elastic modulus in the normal direction and tangential direction respectively. With this, we can calculate the accrued friction force with the equation:
\begin{align}\label{eq:nonholonomic-force}
\dcontactForce{i} = H(-\sdf(\hydroPose{}, \objectPose{}))
\begin{bmatrix}
EA_i & 0  & 0 \\
0  & KA_i & 0 \\
0  & 0  & KA_i 
\end{bmatrix}
\bold R_{ci}(\objectPose{})^\top \dobjectPose{}
\end{align}
where $H$ is the Heaviside step function and $\bold R_{ci}^\top$ rotates $\dobjectPose{}$ to be in the contact frame. We add $H$ to ensure we only accumulate our reaction forces when the object is inside $\mathcal H$. 

Similar to \cite{todorov_implicit} we enforce the Coulomb Friction Law by projecting the contact force $\contactForce{i}$ onto the friction cone:
\begin{align}
\projectcontactForce{i} = \begin{bmatrix}
    \relu(\contactForce{i}_n) \\
    \min(1, \frac{\mu \max(\contactForce{i}_n, 0)} {\|\contactForce{i}_t\| }) \contactForce{i}_t
\end{bmatrix}
\end{align}
We also model a force reset which ensures the forces are only active when the object resides in $\mathcal H$, allowing forces to be reset when contacts transition out from $\mathcal H$.
\begin{align}
\resetcontactForce{i} = H(-\sdf(\hydroPose{}, \objectPose{})) \projectcontactForce{i}
\end{align}

We finally sum the wrench applied by each $\resetcontactForce{}$ and apply them to both the hydroelastic and rigid objects.
\begin{align}
\externalWrench^\text{H} = \sum_{i=1}^{N \times M} {\contactJacobianHydro{}}^\top\resetcontactForce{i} \\
\externalWrench^\text{O} = \sum_{i=1}^{N \times M} {\contactJacobianObj{}}^\top\resetcontactForce{i} \\
\contactForces{} = \begin{bmatrix}
    \resetcontactForce{1} & \resetcontactForce{2} & ... & \resetcontactForce{N\times M}
\end{bmatrix}^\top
\end{align}





We wish to discretize our dynamics and will specifically focus on discretizing $\dcontactForce{}$. The most straightforward discretization using first-order approximation gives the following:
\begin{align*}
    \contactForce{i}_{k+1} = \contactForce{i}_{k} + \dcontactForce{i}_k h
\end{align*} where $h$ is the time-step. 
However, this discretization is not informative for in-contact and out-of-contact transition since it assumes that the rigid object will always be inside or outside of $\mathcal H$ from $k$ to $k+1$. This can introduce errors that can propagate and increase over time. 
To increase the accuracy of our model, we seek a more accurate discretization of in-penetration and out-of-penetration transitions. In particular, we will compute the in-penetration fraction of $\dcontactForce{i,k}$ from  $k$ and $k+1$ using $\sdf{}$.

First, for each differential of area, we compute their displacement in $\mathcal H$, denoted by $\bold d$. This corresponds to the deformation of the hydroelastic element during that step.

\begin{align}
    \alpha_d &= \frac{\relu(-\sdf_{k+1}) - \relu(-\sdf_{k})}{\sdf_k - \sdf_{k+1}} \\
    \bold d &= \alpha_d \bold r =  \alpha_d (\bold p_k - \bold p_{k+1})
\end{align}

where $\sdf_{k} = \sdf{}(\bold p_k)$.
Note that $\alpha_d$ corresponds to the fraction of the hydroelastic element displacement $\bold r$ inside the hydroelastic domain $\mathcal H$ for the . If both $\bold p_k$ and $\bold p_{k+1}$ are outside the hydroelastic element, i.e. $\sdf_{k},\sdf_{k+1} \geq0$, then $\alpha_d =0$. Furthermore, if they are both in $\mathcal H$, i.e. $\sdf_{k},\sdf_{k+1} \leq 0$, then $\alpha_d = 1$.

Next, we project $\bold d$ into normal and tangential components of the contact frame.
$d_n = \langle \bold d, \hat{\bold n}\rangle$ is the projection of $\bold d$ along the normal direction of the contact frame $\hat{\bold n}$.
$\bold d_t = \bold d - d_n \hat{\bold n}$ is the projection of $\bold d$ in the contact plane.
We update the normal forces using the normal displacement $d_n$ and the normal elastic modulus $E$, and tangential forces with tangential displacement $\bold d_t$ and tangential elastic modulus $K$.
\begin{align}
    f_{n,k+1}^{\text{c}_i} &= f_{n,k}^{\text{c}_i} + E A_i d_n \\
    \bold f_{t,k+1}^{\text{c}_i} &= {\bold f}_{t,k}^{\text{c}_i} + K A_i \bold d_t
\end{align}

We then enforce the friction cone constraint. 
\begin{align}
    \bar{f}_{n,k+1}^{\text{c}_i} &= \relu(f^{\text{c}_i}_{n,k+1}) \\
    \bar{\bold f}^{\text{c}_i}_{t,k+1} &= \min(1, \frac{\mu \bar{f}^{\text{c}_i}_{n,k+1}}{\| \bold f^{\text{c}_i}_{t,k+1} \|_2})\bold f^{\text{c}_i}_{t,k+1}
\end{align}




Finally, we calculate force reset if the rigid object exits $\mathcal H$:
\begin{align}
    \resetcontactForce{i}_{k+1} = H(-\phi_{k+1})(\bar{f}^{\text{c}_i}_{n,k+1}\hat{\bold n}+\bar{\bold f}^{\text{c}_i}_{t,k+1})
\end{align}


\subsection{Smoothed Hydroelastic Model}
\label{sec:methods:smoothed_hydro_model}

To enhance the model's computational efficiency and differentiability, we introduce smoothing by relaxing the non-holonomic incremental hydroelastic model. Specifically, we set $\alpha_d = 1$ and replace the Heaviside step function $H$ with a $\texttt{sigmoid}_\beta$ function. This modification weights the force contributions based on the displacement's position relative to the hydroelastic domain $\mathcal{H}$. Displacements outside $\mathcal{H}$ contribute minimally, while those inside the domain contribute fully, ensuring a smooth transition at the boundary. This approach mitigates abrupt changes in force computations making the model suitable for gradient-based optimization.


\begin{figure}
    \centering
    \includegraphics[width=\linewidth]{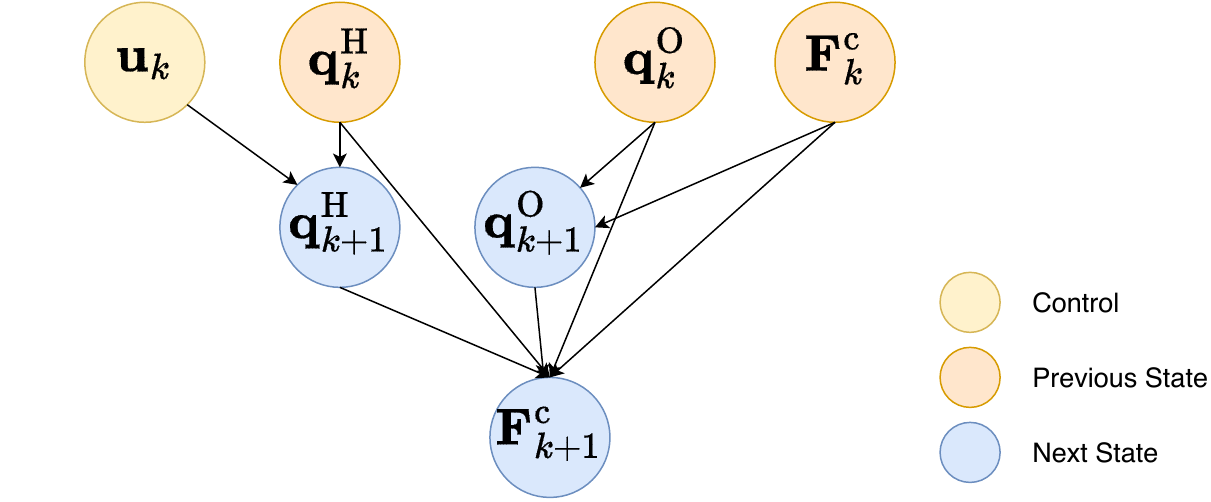}
    \caption{\textbf{Dynamics Computation Graph:} Showing the dependencies to compute next state $\bold x_{k+1}$ from previous state $\bold x_k$ and control $\bold u_k$.}
    \label{fig:hydro_computation_graph}
    \vspace{-15pt}
\end{figure}

\subsection{Quasi-dynamic Formulation}
\label{sec:methods:quasidyn_formulation}

Given the path-dependent hydroelastic model described in section \ref{sec:methods:model}, we can integrate it to recover the dynamics. Given the control actions $\action{}\in \sethree$ as changes of poses of the hydroelastic bodies positions, the dynamics take the following form

\begin{equation*}
    \state{k+1} = f_\text{dyn}(\state{k}, \action{k})
\end{equation*}

where the state is composed by the object pose, hydroelastic bodies pose, and contact forces:

\begin{equation*}
    \state{k} = \begin{bmatrix} \objectPose{k} & \hydroPose{k} & \contactForces{k} \end{bmatrix}^\top
\end{equation*}

We adopt a quasi-dynamic integration scheme, following prior work where it has been shown to be effective for modeling contact-rich interactions \cite{aceituno2020global, pang2023global}. The procedure consists of the following steps, illustrated in Fig.~\ref{fig:hydro_computation_graph}:

\begin{enumerate}
    \item Apply control actions.
        \begin{equation*}
            \hydroPose{k+1} = \hydroPose{k} + \action{k}
        \end{equation*}
    \item Aggregate contact forces
        \begin{equation*}
            \externalWrench^\text{O} = \sum_{i=1}^{\numContacts\times \numHydros} {\contactJacobianObj{i}}^\top \resetcontactForce{i}
        \end{equation*}
    \item Integrate object dynamics
        \begin{equation*}
            \objectPose{k+1} = \objectPose{k+1} + \frac{h^2}{\epsilon} [\inertiaMatrix^{-1}(\externalWrench^\text{O} + \taug)]
        \end{equation*}
    \item Compute next contact forces
        \begin{equation*}
            \contactForces{k+1} = \texttt{compute\_forces}(\objectPose{k+1}, \hydroPose{k+1}, \objectPose{k}, \hydroPose{k}, \contactForces{k})
        \end{equation*}
\end{enumerate}


\subsection{Control Formulation}
\label{sec:methods:control}

In this section, we present the trajectory optimization formulation, which integrates the proposed hydroelastic model with model predictive control (MPC). The objective is to determine a sequence of control actions $\action{} \in \sethree$ that minimizes a prescribed cost function while satisfying the system dynamics governed by $f_\text{dyn}$.


\begin{equation}
\label{eq:mpc_optimization}
\begin{aligned}
\underset{\action{0:K-1}}{\min} \quad 
& J
= \Phi(\state{K}) + \sum_{k=0}^{K-1} \ell_k(\state{k}, \action{k}) \\
\text{s.t.} \quad 
& \state{k+1} = f_{\text{dyn}}(\state{k}, \action{k}), \quad  k=0,\dots,K-1, \\
& \state{k} \in \mathcal{X}_k, \quad \action{k} \in \mathcal{U}_k, \quad k=0,\dots,K-1, \\
& \state{0} = \state{\text{init}}.
\end{aligned}
\end{equation}

Here, $\ell_k(\cdot)$ denotes the stage cost, and $\Phi(\cdot)$ the terminal cost, both assumed to be differentiable.  

Since the dynamics and cost are differentiable, we employ the gradient-based trajectory optimization method from \cite{cem_with_gradients}. This approach integrates Cross Entropy Methods (CEM) with gradient descent, mitigating issues with poor initialization and the locality of purely gradient-based optimization. The procedure is summarized in Algorithm~\ref{alg:gradient_optimization}.  

To explore a broad solution space, we parallelize over $B$ trajectories at each iteration. The best $M$ trajectories (lowest cost) are used to update the control distribution parameters, while the lowest-cost trajectory determines the control applied to the system.  

At each time step, sensor feedback and state estimation provide an updated initial condition, from which the optimization is re-solved. This receding-horizon strategy introduces feedback, enabling adaptation to disturbances, modeling errors, and environmental uncertainty.  


\begin{algorithm}
\caption{Iterative Cross-Entropy Gradient-Based Trajectory Optimization}
\label{alg:gradient_optimization}
\SetKwInOut{Input}{Input}
\SetKwInOut{Output}{Output}
\Input{
Reference distribution parameters $\mu_{0:K-1}, \Sigma_{0:K-1}$, 
dynamics $f$, 
cost function $J$, 
number of parallel trajectories $B$, 
gradient descent iterations $G$,
elite set size $M \leq B$.
}
\Output{
Optimized control sequence $\mathbf{u}_{0:K-1}^*$.
}

\For{each time step $h=0$ to $H-1$}{
    Measure current state $\state{h}$\;
    Sample $B$ control sequences $\{\mathbf{u}_{h:K-1}^{(b)}\}_{b=1}^B$ from $\mathcal N(\mu_{h:K-1}, \Sigma_{h:K-1})$\;
    \For{each gradient iteration $i=1$ to $G$}{
        \For{each trajectory $b = 1$ to $B$}{
            Roll out trajectory $\{\mathbf{x}_k^{(b)}\}_{k=h}^K$ with dynamics $\mathbf{x}_{k+1}^{(b)} = f(\mathbf{x}_k^{(b)}, \mathbf{u}_k^{(b)})$\;
            Compute cost $J^{(b)}$\;
            Compute gradient $\nabla_{\mathbf{u}_{h:K-1}^{(b)}} J^{(b)}$\;
            Update $\mathbf{u}_{h:K-1}^{(b)}$ with gradient descent\;
        }
    }
    Select top $M$ trajectories with lowest cost and update $\mu_{h:K-1}, \Sigma_{h:K-1}$\;
    Apply the first control ${\action{h}}$ from the best trajectory to the system\;
    Resample the bottom $B-M$ trajectories from $\mathcal N(\mu_{h:K-1}, \Sigma_{h:K-1})$\;
}
\end{algorithm}




%

\begin{figure*}
    \centering
    \includegraphics[width=\linewidth]{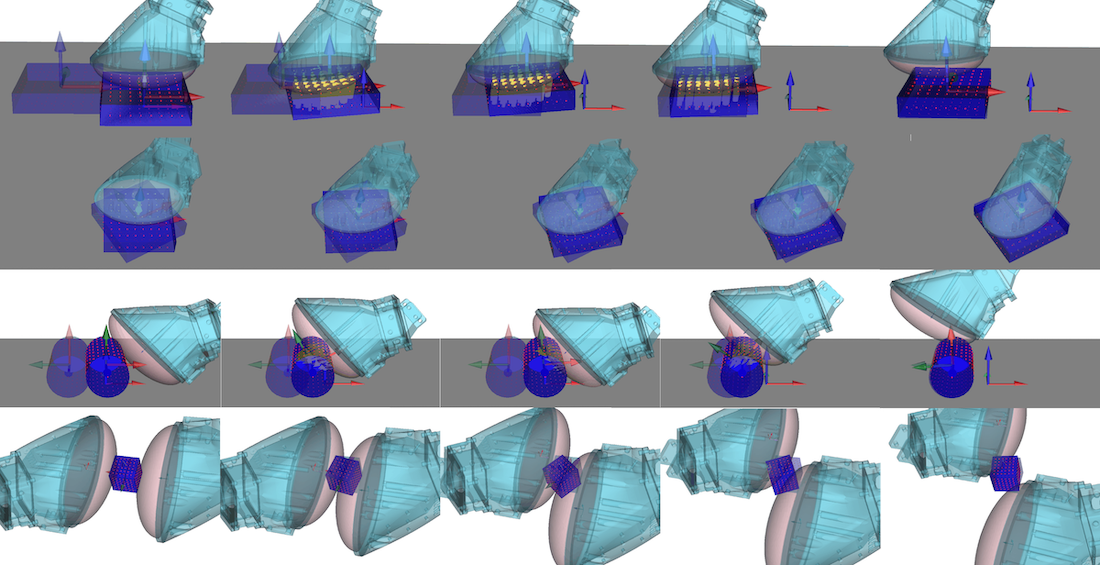}
    \vspace{-15pt}
    \caption{\textbf{Simulation}: 
    Sequential frames illustrating the four simulated manipulation tasks: planar pushing (top row), planar rotation (second row), rolling (third row), and in-hand rotation (bottom row). Note that both the bubble (pink) are modeled as hydroelastic elements and the object (solid blue) is treated as rigid. Contact forces are visualized in yellow, and goals are shown with transparency (reduced alpha).}
    \label{fig:sim_vis}
    \vspace{-5pt}
\end{figure*}

\section{Experiments and Results}
\label{sec:experiments}

\begin{figure}
    \centering
    \includegraphics[width=0.8\linewidth]{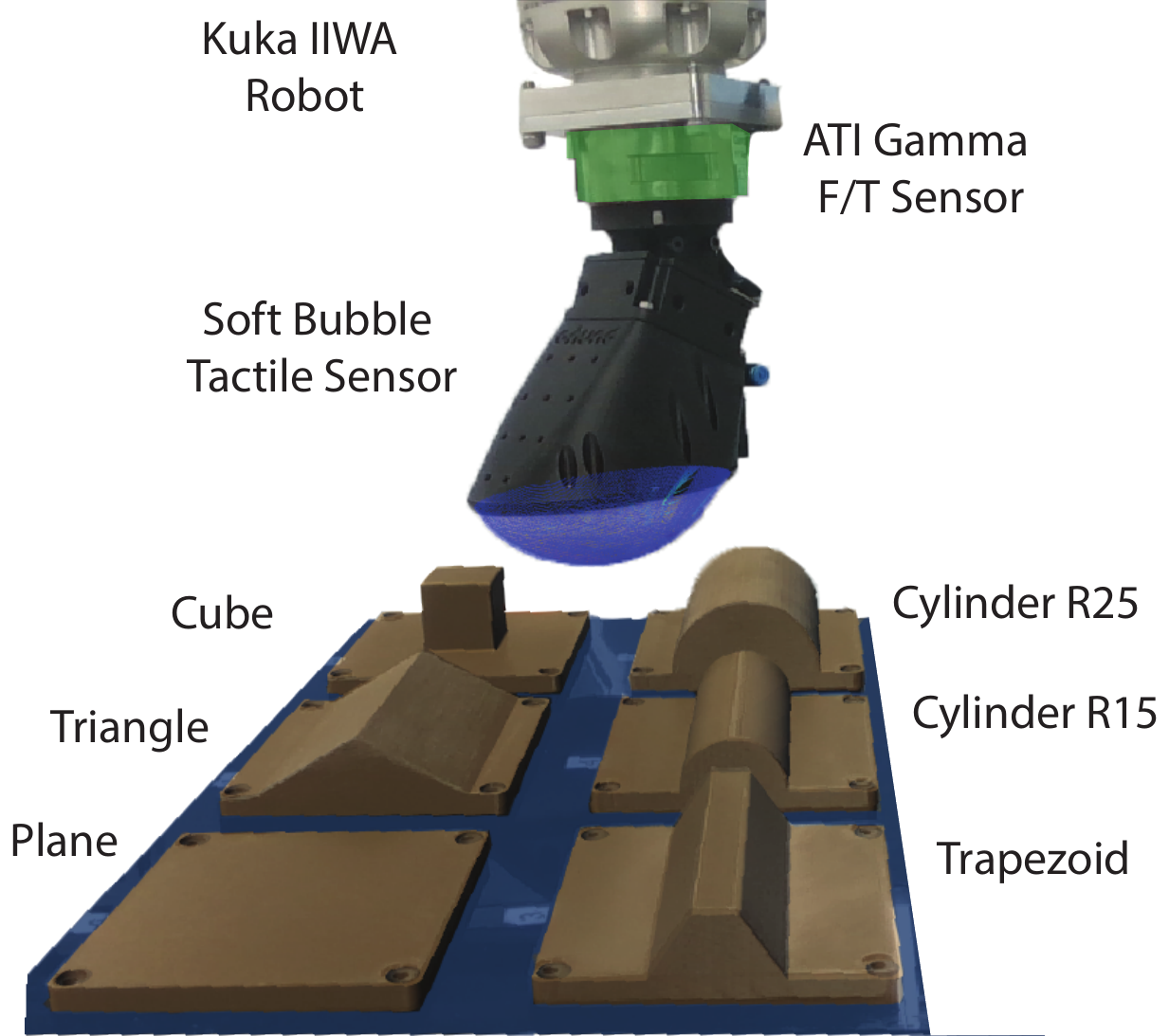}
    \caption{\textbf{Wrench Transmission Evaluation Setup}: The robot equipped with F/T and Soft Bubble sensors pressing a set of geometries. }
    \label{fig:wrench_setup}
    \vspace{-15pt}
\end{figure}

\subsection{Tactile Feedback}
We evaluate our method by modeling the dynamics of the Soft Bubbles tactile sensor \cite{soft_bubbles} for contact-rich robotic manipulation tasks. This sensor provides high-resolution information about deformations of a highly compliant, pressurized latex membrane. An internal camera captures changes in the membrane geometry, enabling detailed observation of contact interactions. In this work, we utilize the tactile sensor data as feedback to estimate contact forces. By combining robot proprioception with the sensor data, we obtain the pose of each hydroelastic object $\hydroPose{}$, partial information about the object pose $\objectPose{}$, and the membrane deformations, which are used to compute the contact forces $\contactForces{}$.

Specifically, to recover the contact forces $\contactForces{}$ from tactile observations, we use the sensed object pose and the hydroelastic object pose to determine the membrane deformation $\bold{d}_i$ for each discretized point $\bold{p}_i$ on the object surface. We then apply the compliance model described in Section \ref{sec:methods:model} to compute the forces as follows:
\begin{align}
    \bold{d}_{n,i} &= \langle \bold{d}_i, \hat{\bold{n}}\rangle \hat{\bold{n}}, \\
    \bold{d}_{t,i} &= \bold{d}_i - \bold{d}_{n,i}, \\
    {\contactForce{}}_{i} &= EA_i \bold{d}_{n,i} + KA\bold{d}_{t,i}.
\end{align}

While this approach relies on a linear approximation of the sensor compliance, it effectively corrects for dynamical mismatches and accounts for sliding drift mismatch. Also, this regularizes the smoothed dynamics and ensures a more accurate representation of the contact forces during manipulation tasks.




\subsection{Hydroelastic Baselines}
To evaluate the performance of our proposed non-holonomic (NH) model, we compare against established holonomic hydroelastic models. Specifically, we evaluate two baselines: (i) the \emph{Pressure Field} (PF) model \cite{pressure_field}, which estimates contact interactions by integrating the full pressure distribution over the contact patch, including both normal and tangential components of force; and (ii) the \emph{Pressure Field Frictionless} (PFF) model \cite{pressure_field}, an ablated variant of PF in which the tangential components are set to zero, thereby restricting the model to purely normal force transmission. These baselines enable us to isolate the contribution of tangential shear forces and highlight the advantages of our non-holonomic formulation in capturing realistic force transmission during contact-rich manipulation.

\begin{figure*}
    \centering
    \includegraphics[width=\linewidth]{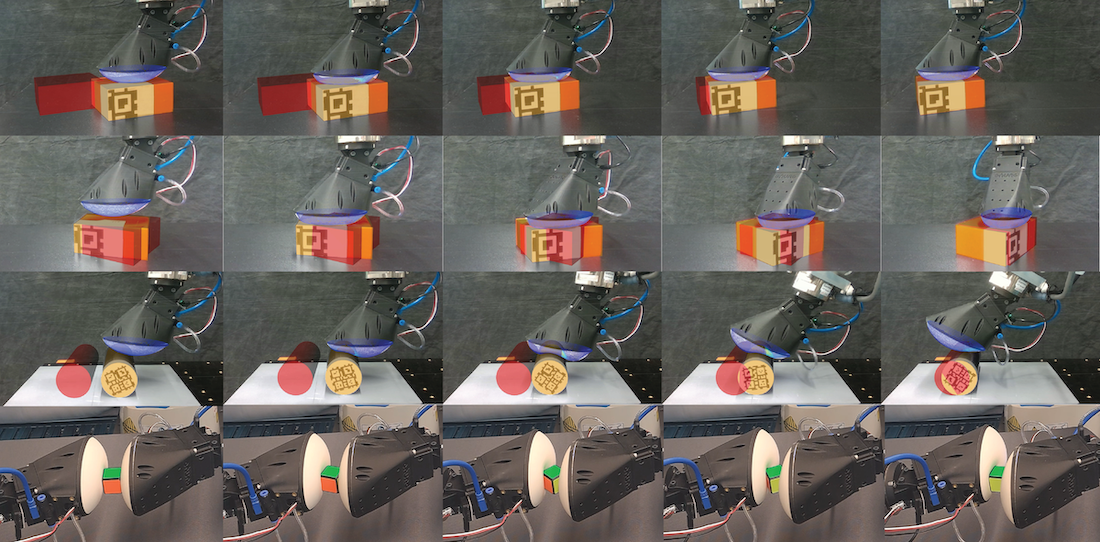}
    \vspace{-15 pt}
    \caption{\textbf{Real-world experiments.} Sequential frames illustrating the four manipulation tasks: planar pushing (top row), planar rotation (second row), rolling (third row), and in-hand rotation (bottom row). The orange outline indicates the object pose, and the red outline denotes the goal pose.}
    \label{fig:rw_vis}
\end{figure*}
 
\subsection{Wrench Transmission Benchmark}

\begin{table}[h!]
\vspace{-5pt}
\centering
\begin{tabular}{lcccccc}
\toprule
\multirow{4}{*}{Object} & \multicolumn{6}{c}{Wrench RMSE [N] ($\downarrow$)} \\
\cmidrule(lr){2-7} & \multicolumn{2}{c}{NH (Ours)} & \multicolumn{2}{c}{PF \small (Baseline)} & \multicolumn{2}{c}{PFF \small (Baseline)} \\
\cmidrule(lr){2-3} \cmidrule(lr){4-5} \cmidrule(lr){6-7}
 & \dc Mean & Std  &  \dc Mean & Std  & \dc Mean  &  Std \\ \midrule

Plane &
\dc \textbf{0.43} & 0.12 &
\dc 2.32 & 0.66 &
\dc 1.25 & 0.68 \\

Triangle  &
\dc \textbf{0.44} & 0.22 &
\dc 1.31 & 0.40 &
\dc 0.97 & 0.38  \\

Trapezoid  &
\dc \textbf{0.80} & 0.28 &
\dc 1.65 & 0.71 &
\dc 1.64 & 0.55  \\

Cube &
\dc \textbf{0.57} & 0.12 &
\dc 0.87 & 0.20 &
\dc 0.69 & 0.23 \\

Cylinder R15 &
\dc 1.80 & 0.50 &
\dc 2.64 & 0.72 &
\dc \textbf{1.58} & 0.44 \\

Cylinder R25 &
\dc \textbf{1.69} & 0.37 &
\dc 3.20 & 0.85 &
\dc 1.75 & 0.57 \\

\bottomrule
\end{tabular}
\vspace*{-3pt}
\caption{\textbf{Hydroelastic Model Benchmark:} We collect 40 trajectories per object, each being up to 40 steps. We compute the root mean squared error (RMSE) between the ground truth wrenches, measured at the robot wrist, and the model predictions. We report the mean (highlighted) and standard deviation.
}
\label{tab:model_benchmark}
\vspace*{-20pt}
\end{table}











We evaluate the accuracy of predicted force transmission from our model against baseline methods across a diverse set of real-world contact interactions.
Figure \ref{fig:wrench_setup} visualizes the data-collection setup, where the robot is equipped with the Soft Bubble tactile sensor and makes contact with a variety of object geometries. We assess how well each model predicts the total transmitted wrench, which is measured by a Gamma ATI force-torque (F/T) sensor mounted on the robot wrist.

Table~\ref{tab:sim_real_evaluation} reports the root-mean-squared error (RMSE) of the predicted wrench relative to the ground-truth F/T measurements. Our results indicate that the non-holonomic model provides the most accurate characterization of the transmitted wrench, outperforming all baselines. Interestingly, the PFF baseline — despite its simplicity compared to PF — yields better wrench predictions than PF, highlighting the limitations of tangential force modeling in the PF formulation for compliant elements.







\subsection{Contact Patch Regulation for Manipulation}

\begin{table*}[h!]
\centering
\begin{tabular}{lcccccccccccc}
\toprule
\multirow{4}{*}{Task} & \multicolumn{4}{c}{Non-holonomic (Ours)} & \multicolumn{4}{c}{Pressure Field (Baseline)} & \multicolumn{4}{c}{PF Frictionless (Baseline)} \\
\cmidrule(lr){2-5}\cmidrule(lr){6-9}\cmidrule(lr){10-13}
 & \multicolumn{2}{c}{OL Error [mm]} & \multicolumn{2}{c}{CL Error [mm]} 
 & \multicolumn{2}{c}{OL Error [mm]} & \multicolumn{2}{c}{CL Error [mm]}
 & \multicolumn{2}{c}{OL Error [mm]} & \multicolumn{2}{c}{CL Error [mm]} \\
\cmidrule(lr){2-3}\cmidrule(lr){4-5}
\cmidrule(lr){6-7}\cmidrule(lr){8-9}
\cmidrule(lr){10-11}\cmidrule(lr){12-13}
 & \dc Mean $\downarrow$ & Std $\downarrow$ & \dc Mean $\downarrow$ & Std $\downarrow$
 & \dc Mean $\downarrow$ & Std $\downarrow$ & \dc Mean $\downarrow$ & Std $\downarrow$ 
 & \dc Mean $\downarrow$ & Std $\downarrow$ & \dc Mean $\downarrow$ & Std $\downarrow$ \\ \midrule

\multicolumn{13}{l}{\textbf{Simulation}} \\ \midrule
Planar Pushing    & \dc 2.21  & 0.68 & \dc \textbf{0.02} & 0.01 & \dc2.79  & 0.74 & \dc2.54  & 1.59  & \dc101   & 2.07  & \dc100   & 2.87  \\
Planar Rotation   & \dc3.20  & 0.88 & \dc \textbf{0.04} & 0.05 & \dc32.6  & 17.9  & \dc20.1  & 22.3  & \dc40.7  & 10.1  & \dc40.6  & 12.9  \\
Rolling           & \dc \textbf{2.24}  & 1.83  & \dc 6.14   & 5.56   & \dc 41.5  & 24.0  & \dc 42.8  & 26.8  & \dc 47.3  & 18.4  & \dc 49.1  & 18.8  \\
In-hand Rotation  & \dc3.98  & 1.39  & \dc \textbf{0.66}  & 1.03   & \dc8.94  & 2.91  & \dc7.37  & 3.34  & \dc9.54  & 3.77  & \dc9.58  & 3.79  \\ \midrule

\multicolumn{13}{l}{\textbf{Real World}} \\ \midrule
Planar Pushing    & \dc 6.59  & 3.62  & \dc \textbf{2.09}   & 1.02   & \dc 58.5  & 9.63  & \dc45.3  & 16.7  & \dc 73.3  & 28.4  & \dc 90.6  & 20.2  \\
Planar Rotation   & \dc 14.9  & 9.34  & \dc \textbf{3.87}   & 1.23   & \dc 57.5  & 17.3  & \dc57.8  & 16.5  & \dc 37.9  & 7.25  & \dc 37.9  & 7.28  \\
Rolling           & \dc 21.2  & 15.0  & \dc \textbf{9.02}   & 2.86   & \dc 61.1  & 18.0  & \dc53.1  & 26.9  & \dc 52.8  & 19.6  & \dc 62.5  & 23.1  \\
In-hand Rotation  & \dc5.19  & 0.84 & \dc \textbf{4.07}   & 0.89  & \dc 12.7  & 4.75  & \dc9.14  & 4.78  & \dc14.3  & 8.34  & \dc10.8  & 4.67  \\

\bottomrule
\end{tabular}
\vspace*{-2pt}
\caption{\textbf{Optimized Trajectory Control Evaluation:} Benchmark of different models (Non-holonomic, PF, PF Frictionless) in both simulation and real-world experiments. Each cell reports Mean and Std for open-loop (OL) and closed-loop (CL) pose tracking errors [mm]. Mean columns are highlighted.}
\label{tab:sim_real_evaluation}
\vspace*{-25pt}
\end{table*}

In this section, we evaluate the effectiveness of our model for contact-rich manipulation tasks, focusing on its integration with gradient-based control. 
We consider tasks that require active regulation of the contact patch and modulation of force transmission through shear. Specifically, we study:  
i) planar pushing,  
ii) planar rotation,  
iii) cylindrical rolling, and  
iv) bi-manual in-hand rotation.  
These tasks are evaluated both in quasi-dynamic simulation and on real-world robotic platforms. Figures ~\ref{fig:sim_vis} and \ref{fig:rw_vis} illustrate the simulation and real-world environments, respectively. 
For each task, the robot moves the object towards 10 randomly sampled goal configurations, repeated per method to be evaluated. For real-world execution, we use AprilTags to obtain ground-truth object poses.

Table~\ref{tab:sim_real_evaluation} summarizes the quantitative evaluation results across simulation and real-world settings.  
The PFF baseline consistently fails to induce meaningful object motion in planar pushing, rotation, and rolling, both in simulation and in the real world, due to its inability to generate shear forces at the contact patch. In contrast, the PF baseline succeeds in simulation but exhibits unrealistic behaviors that do not transfer to real-world settings—for instance, in planar rotation and rolling tasks, it occasionally moves the object away from the goal. Overall, our proposed method outperforms both baselines, reducing errors across all tasks. While the real-world execution error remains higher than in simulation, our approach narrows the sim-to-real gap more effectively than competing methods. As expected, closing the control loop improves performance compared to open-loop execution. 

\section{Discussion, Limitations, and Future Work}
\label{sec:discussion}
The proposed non-holonomic hydroelastic model, combined with tactile feedback, shows strong potential for accurately estimating contact forces and compensating for dynamical mismatches in contact-rich manipulation. Nevertheless, several limitations remain, pointing to avenues for future work.
A key limitation lies in the state representation: the explicit inclusion of contact forces causes the state dimension to scale poorly with object discretization and the number of objects. This results in substantial computational overhead, particularly for methods such as iterative Model Predictive Control (iMPC), where the computation of large Jacobian matrices for locally linearized dynamics becomes prohibitive. Moreover, our method currently applies only in closed-loop scenarios where contact forces can be directly observed or reliably estimated, for example via tactile sensing. Integrating recent approaches for contact force estimation \cite{van2025estimating} \cite{nisp} could help alleviate this restriction and broaden applicability.
Finally, while in this work we primarily evaluate the model within a trajectory optimization setting, future research could explore its use in reinforcement learning. Given its ability to capture the dynamics of highly compliant contacts with relatively low computational overhead, the model may serve as a useful tool for training policies in simulation that transfer effectively to real-world systems.

\addtolength{\textheight}{-570pt}   







\bibliographystyle{ieeetr}
\bibliography{references}

\end{document}